\documentclass[conference,9pt]{IEEEtran}
\IEEEoverridecommandlockouts

\usepackage{cite}
\usepackage{amsmath,amssymb,amsfonts}
\usepackage{algorithmic}
\usepackage{graphicx}
\usepackage{textcomp}
\usepackage{xcolor}
\usepackage{multirow}
\usepackage{cite}
\usepackage{comment}
\usepackage{bm}
\newcommand{\argmax}{\mathop{\rm arg~max}\limits}

\def\BibTeX{{\rm B\kern-.05em{\sc i\kern-.025em b}\kern-.08em
    T\kern-.1667em\lower.7ex\hbox{E}\kern-.125emX}}
\begin{document}

\title{Hypothesis Clustering and Merging: \\Novel MultiTalker Speech Recognition with Speaker Tokens}


\author{\IEEEauthorblockN{Yosuke Kashiwagi\IEEEauthorrefmark{1}, Hayato Futami\IEEEauthorrefmark{1}, Emiru Tsunoo\IEEEauthorrefmark{1}, Siddhant Arora\IEEEauthorrefmark{2} and Shinji Watanabe\IEEEauthorrefmark{2}}
\IEEEauthorblockA{
\IEEEauthorrefmark{1}Sony Group Corporation \\
\{yosuke.kashiwagi,hayato.futami,emiru.tsunoo\}@sony.com
}
\IEEEauthorblockA{
\IEEEauthorrefmark{2}Carnegie Mellon University \\
\{siddhana,swatanab\}@andrew.cmu.edu
}
}

\maketitle

\begin{abstract}
In many real-world scenarios, such as meetings, multiple speakers are present with an unknown number of participants, and their utterances often overlap.
We address these multi-speaker challenges by a novel attention-based encoder-decoder method augmented with special speaker class tokens obtained by speaker clustering. 
During inference, we select multiple recognition hypotheses conditioned on predicted speaker cluster tokens, and these hypotheses are merged by agglomerative hierarchical clustering (AHC) based on the normalized edit distance. 
The clustered hypotheses result in the multi-speaker transcriptions with the appropriate number of speakers determined by AHC.
Our experiments on the LibriMix dataset demonstrate that our proposed method was particularly effective in complex 3-mix environments, achieving a 55\% relative error reduction on clean data and a 36\% relative error reduction on noisy data compared with conventional serialized output training.

\end{abstract}


\begin{figure*}[t]
  \centering
  \includegraphics[width=0.95\linewidth]{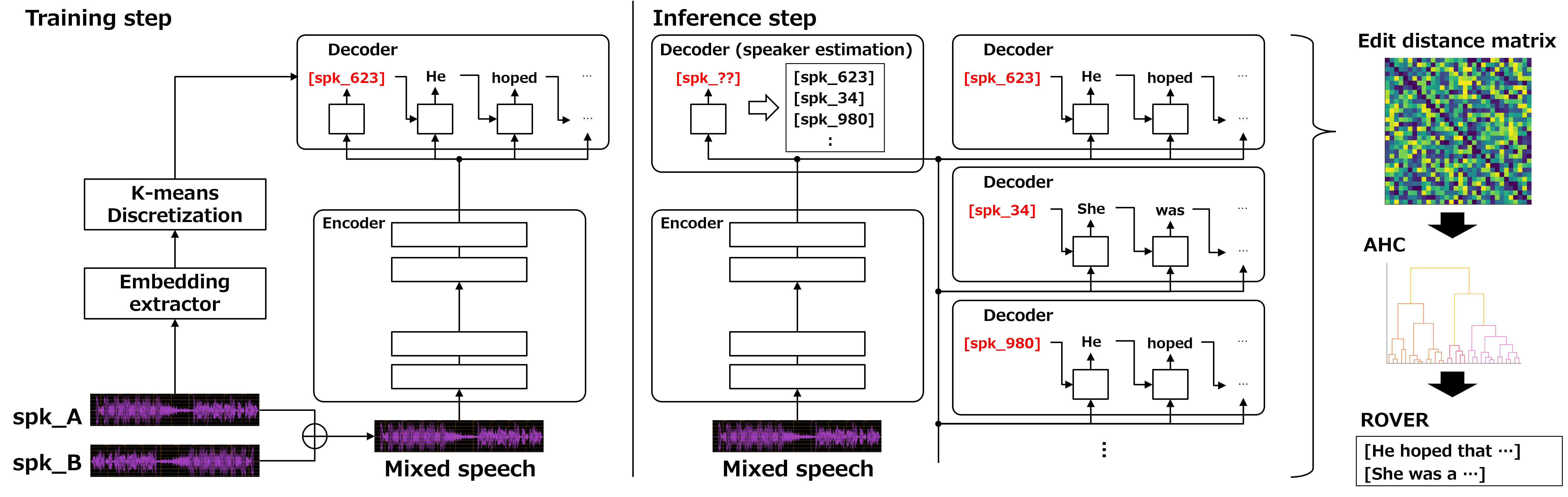}
\caption{Training and inference procedures of hypothesis clustering and merging.}
  \label{fig:multilingualASR}
\end{figure*}

\section{Introduction}
Overlapped speech recognition has recently garnered significant attention due to its increasing demand in various real-world applications \cite{haeb2019speech}.
There are two primary approaches to handling overlapped speech in multi-talker automatic speech recognition (MT-ASR) \cite{masumura2024unified}.
One is target-speaker MT-ASR, where the speech of a predefined speaker is transcribed using prior information on the specific speaker \cite{kanda2019auxiliary,kanda2021investigation,kanda2021endtoend,ma2024extending}.
Another approach is target-speaker-free MT-ASR including permutation invariant training (PIT) \cite{yu2017permutation,seki2018purely} and serialized output training (SOT) \cite{kanda2020serialized}, which transcribe all speakers' speech without relying on speaker-specific information.

Target-speaker MT-ASR uses speaker embeddings, such as x-vector \cite{snyder2018x}, to represent target speaker information and use them.
Recent advancements in powerful speaker representation models such as ECAPA-TDNN \cite{desplanques2020ecapa} and TitaNet \cite{koluguri2022titanet} have significantly enhanced the performance on MT-ASR \cite{zhang2023conformer}.
However, the approach requires prior speaker enrollment, which is often costly to prepare and limits its range of applicable scenarios.

On the other hand, target-speaker-free MT-ASR, PIT and SOT, do not requires the speaker enrollment and can transcribe overlapped speech without speaker-specific information. 
PIT allows simultaneous recognition of multiple speakers by training the model to be permutation invariant \cite{yu2017permutation,seki2018purely}.
This ensures that the output corresponds to the correct speaker, regardless of the order in which the speakers are processed.
This makes PIT particularly effective in scenarios with a fixed number of speakers.
SOT further improves this by enabling a sequentially concatenated transcription of overlapped speech using a single output layer, which reduces the complexity of handling speaker permutations.
Consequently, SOT autoregressively predicts a concatenated multi-speaker transcription more stably and coherently by considering the dependencies between the transcriptions of each speaker \cite{kanda2020serialized}.

However, both PIT and SOT are less flexible when dealing with varying numbers of speakers.
The effectiveness of PIT decreases as the number of speakers increases due to the exponential growth in the number of possible speaker permutations considered during training. 
Similarly, SOT's performance can be degrade by a larger number of speakers due to very long serialized sequence.
Also, because SOT does not explicitly constrain the transcriptions of different speakers to be distinct, it sometimes outputs the same transcription repeatedly.

In this study, we propose a novel method called Hypothesis Clustering and Merging (HCM) for overlapped speech recognition, focusing on target-speaker-free MT-ASR.
HCM first defines speaker clusters by clustering speaker embeddings using \(k\)-means clustering.
This was inspired by the process of discretizing SSL embeddings so that they can be predicted by language models \cite{borsos2023audiolm,rubenstein2023audiopalm,shon2024discreteslu,yang2024towards,chang2023exploration}.
This technique was also used in previous study to estimate speaker embeddings from overlapped speech \cite{makishima2024somsred}.
Subsequently, each speaker cluster ID is treated as a special token and prepended to the beginning of the transcription during training as a prompt.
A standard attention-based encoder-decoder (AED) is used.
During decoding, multiple speaker tokens are hypothesized from the speaker token's probability estimated from the mixed audio and used as prompts to generate the corresponding multiple transcriptions.
The decoding is carried out under the assumption that speakers of the candidate clusters exist in the overlapped speech.
Finally, multiple transcription hypotheses corresponding to the candidate speaker tokens are clustered using Agglomerative Hierarchical Clustering (AHC) \cite{lance1967general}, based on the normalized edit distances between the hypothesized transcriptions.
The multiple transcriptions in a single hypothesis cluster are then merged using the Recognizer Output Voting Error Reduction (ROVER) technique \cite{fiscus1997post}, yielding the final set of transcriptions.
The number of speakers is determined simply by applying a threshold to the AHC process.

HCM has several advantages.
Unlike PIT, it scales to a large number of speakers without exponentially increasing training costs.
Unlike SOT, HCM does not have to predict a long sequence, and can avoid to generate the same text repeatedly via the merging process.
HCM can explicitly constrain the output transcription to differ by speaker.
We verified the effectiveness of the proposed method through experiments using Librimix data \cite{cosentino2020librimix}.
In particular, in the three-speaker setting, we achieved a 55\% relative error reduction on clean data and a 36\% relative error reduction on noisy data compared to SOT.

\section{Related works}

\subsection{Permutation invariant training}

PIT is a crucial technique in speech separation \cite{yu2017permutation,seki2018purely,kolbaek2017multitalker,xu2018single,qian2018single}. 
In traditional speech separation, there is ambiguity in determining which estimated output corresponds to the correct reference signal, a challenge referred to as the "permutation problem." 
To address this, permutation-invariant training (PIT) is used, where the model evaluates all possible output-reference pairings and selects the optimal match to minimize the loss function.
The loss function can be flexibly chosen depending on the task.
For example, in speech separation, the loss might be calculated in the spectral domain, while in overlapped speech recognition, the cross-entropy loss is typically used.
PIT enables the model to be trained without being affected by the correspondence of speakers, allowing to perform permutation-invariant speech separation.

However, while PIT performs well when there are small numbers of speakers, the number of possible pairings increases exponentially as the number of speakers grows.
This leads to a exponential increase in the computational cost during training.
This issue becomes even more pronounced in recent PIT-based speech recognition models, where not only the decoder but also the encoder branches off in the middle.
On the other hand, our approach explicitly associates each speaker token with its corresponding transcription during data preparation, thereby eliminating permutation ambiguity.


\subsection{Serialized output training}



SOT is based on an AED and, unlike traditional PIT, it generates transcriptions of overlapped speech sequentially using a single output layer \cite{kanda2020serialized,kanda2022streaming,papi2023token}. 
Specifically, SOT introduces a special symbol, \(\langle sc \rangle\), which refers "speaker change," and is used to separate the speech of different speakers, thereby allowing the model to transcribe the speech in a sequential manner.

One of the advantage of SOT is its efficiency with reduced computational cost compared with PIT by fixing a speaker order based on a First-In, First-Out (FIFO) approach.
Another advantage is that the "speaker change" token allows the model to operate independently of the number of speakers.
This token enables the model to implicitly capture the dependencies between the transcriptions of multiple speakers.
SOT cannot explicitly correspond the individual transcriptions to the speaker because it is trained to implicitly handle transcriptions only in the serialized form based on the training data.
As the number of speakers increases, the sequence length that the decoder must handle grows longer, making it more difficult to manage these dependencies. 
Consequently, there can be cases where the same transcription is mistakenly repeated.

Additionally, the sequence length of the output increases proportionally with the number of speakers, which can lead to issues in computational cost and instability in generating hypotheses during inference.

\section{Proposed method}

\subsection{Speaker clustering}
\label{ssec:speaker_clustering}

Fig. \ref{fig:multilingualASR} provides an overview of HCM.
We first prepare speaker class tokens for HCM. 
We assume that strict identification of each speaker is unnecessary for recognizing overlapped speech \cite{makishima2024somsred}.
Therefore, we cluster speaker embeddings obtained from TitaNet-large \cite{koluguri2022titanet} into \(k\) discrete classes using \(k\)-means clustering.
This clustering scheme is similar to the process of discretizing SSL embeddings so that they can be modeled with language models \cite{borsos2023audiolm,rubenstein2023audiopalm,shon2024discreteslu,yang2024towards,chang2023exploration}.
Although the original speaker IDs are already discrete, clustering them again through embeddings allows us to assign a speaker token even to unknown speakers.

Subsequently, we prepare speaker tokens for the speakers involved in each overlapped speech segment in the training data.
If the overlapped speech is generated through simulation, these tokens can be obtained using the original speech.
Alternatively, the speaker token can be predicted using the single-speaker region from the same speaker as a reference.
This process creates a set of overlapped speech, the text, and the corresponding speaker token.
In the case of 2-mix simulation data, the mixed speech is generated from two original speech sets.
Let \(\mathbf{D}^{\mbox{single}}_1 = \{\mathbf{s}_1, \mathbf{t}_1\}\) and \(\mathbf{D}^{\mbox{single}}_2 = \{\mathbf{s}_2, \mathbf{t}_2\}\) represent the two sets of original speech and corresponding text.
The mixed speech \(\mathbf{s}_{\text{mix}}\) is generated by combining \(\mathbf{s}_1\) and \(\mathbf{s}_2\):
\begin{align}
\mathbf{s}_{\text{mix}} = \alpha_1 \mathbf{s}_1 + \alpha_2 \mathbf{s}_2,
\end{align}
where \(\alpha_1\) and \(\alpha_2\) are the randomly selected mixing weights.
Speaker tokens are calculated from each speech signal using an embedding extractor and \(k\)-means clustering. Let \(\mathbf{e}_1\) and \(\mathbf{e}_2\) be the speaker embeddings extracted from \(\mathbf{s}_1\) and \(\mathbf{s}_2\), respectively. Using \(k\)-means clustering, the corresponding speaker tokens \(c_1\) and \(c_2\) are computed as follows:
\begin{align}
c_1 = \text{k-means}(\mathbf{e}_1), \quad c_2 = \text{k-means}(\mathbf{e}_2).
\end{align}
The resulting set includes the mixed speech \(\mathbf{s}_{\text{mix}}\), the two speaker tokens \(c_1\) and \(c_2\), and the corresponding texts \(\mathbf{t}_1\) and \(\mathbf{t}_2\):
\begin{align}
\mathbf{D}^{\mbox{mix}}_{1,2} = \{\mathbf{s}_{\text{mix}}, (c_1, \mathbf{t}_1), (c_2, \mathbf{t}_2)\}.
\end{align}
Note that we only use the speaker embedding vector in this speaker token preparation stage.
We don't use it for the following training and inference stages, unlike target speaker ASR \cite{kanda2019auxiliary} or speaker-embedding-based SOT \cite{fan2024sa}.

\subsection{Enumerating transcription hypotheses}
\label{ssec:enumerating_hypotheses}

During training (left side in Fig. \ref{fig:multilingualASR}), we perform end-to-end training targeting a label sequence \(\mathbf{y}\), where the speaker token is prepended at the beginning of the transcription:
\begin{align}
\mathbf{y} &= [c, \mathbf{t}^\top]^\top.
\end{align}
This approach is similar to the use of language tokens in multilingual speech recognition \cite{watanabe2017language,toshniwal2018multilingual,radford2023robust}.

During inference (right side in Fig. \ref{fig:multilingualASR}), we first calculate the probabilities of speaker tokens \(P(c | \mathbf{s}_{\text{mix}})\) from the input overlapped speech \(\mathbf{s}_{\text{mix}}\).
We then select the top-\(N\) speaker token candidates \(\{\hat{c}_{1}, \hat{c}_{2}, \dots, \hat{c}_{N}\}\) with the highest probabilities:
\begin{align}
\{\hat{c}_{1}, \hat{c}_{2}, \dots, \hat{c}_{N}\} = \text{Top-N} \left(P(c | s_{\text{mix}})\right).
\end{align}
Since the speaker token appears at the beginning of the transcription, we refer to the initial occurrence probability in the decoder. 
Alternatively, if the model is jointly trained with Connectionist Temporal Classification (CTC) and an attention decoder \cite{hori2017joint}, it is possible to estimate probabilities using the CTC branch.
Subsequently, each candidate speaker token \(\hat{c}_{n}\) is used as a prompt for the decoder, yielding \(N\) recognition results \(\hat{\mathbf{Y}} = \{\hat{\mathbf{y}}_{1}, \hat{\mathbf{y}}_{2}, \dots, \hat{\mathbf{y}}_{N}\}\).
\(\hat{\mathbf{Y}}\) often contains many similar recognition results because the model tends to assign high probabilities to multiple speaker classes that acoustically resemble the actual single speaker.
We explore two approaches to merging these recognition results: a simple voting and our HCM approach.

\subsection{Simple voting}
\label{ssec:simple_voting}

As mentioned earlier, the recognition results often include many similar text segments. 
In this method, we count the occurrences of each candidate transcription \(\mathbf{t}'\) from the set of hypotheses \(\hat{\mathbf{Y}}\). The count is computed as:
\begin{align}
\text{Vote}(\mathbf{t}') = \sum_{n=1}^{N} \mathbf{1}\left(\hat{\mathbf{t}}_{n} = \mathbf{t}'\right),
\end{align}
where \(\mathbf{1}(\cdot)\) is the indicator function that returns 1 if the condition is true and 0 otherwise.
We then select the most frequent transcriptions up to the number of speakers \(K\) present in the overlapped speech:
\begin{align}
\{\mathbf{t}'_{1}, \mathbf{t}'_{2}, \dots, \mathbf{t}'_{K}\} = \text{Top-K} \left( \text{Vote}(\mathbf{t}') \right).
\end{align}
It is important to note that this voting method requires prior knowledge of the number of speakers \(K\).
This simple voting provides a straightforward baseline for comparison with our HCM, which does not require prior knowledge of the number of speakers and can provide more complex merging strategy.

\subsection{Hypothesis clustering and merging}
\label{ssec:HCM}

Beam search is a common method for handling recognition hypotheses.
However, differences in word sequences often have a more significant impact on the score than differences in speaker tokens. 
As a result, hypotheses with different speaker tokens are often pruned earlier in the beam search process, leaving only one recognition result associated with a single speaker token, even in the presence of multiple speakers.
One key idea of HCM is to \textit{redundantly generate more speaker tokens and corresponding recognition results than the actual number of speakers}, rather than attempting to predict the exact number of speakers.

\subsubsection{Clustering hypotheses}
\label{sssec:clustering_hypotheses}

First, the set of recognition results \(\hat{\mathbf{Y}}\) is clustered into \(H\) clusters using text-based normalized edit distance without special speaker tokens.
The normalized edit distance \(\bar{d}_{\text{edit}}(\mathbf{t}_i, \mathbf{t}_j)\) is defined as:
\begin{align}
\bar{d}_{\text{edit}}(\mathbf{t}_i, \mathbf{t}_j) = \frac{d_\text{edit}(\mathbf{t}_i, \mathbf{t}_j)}{\max(|\mathbf{t}_i|, |\mathbf{t}_j|)}.
\end{align}
One of the advantages is that this approach is applicable even when the number of speakers is unknown.
If the number of speakers is known, fixed-cluster clustering can be employed. However, generally, the number of speakers is unknown. 
AHC \cite{lance1967general} can be used to robustly determine the number of speaker clusters through simple thresholding.
AHC is often used for clustering speaker embeddings in diarization tasks \cite{landini2022bayesian}. 
Note that our proposed method uses the normalized edit distance as a distance metric, but it can be easily extended to incorporate other metrics, such as those based on speaker embeddings.

\subsubsection{Merging using ROVER}
\label{sssec:merging_using_rover}

ROVER \cite{fiscus1997post} can utilize information from text segments that are not completely identical.
For each cluster, the text sequences are merged using ROVER. 
First, the subset of \( M_h \) recognition hypotheses \(\hat{\mathbf{Y}}_{h} = \{\hat{\mathbf{y}}_{h,1}, \hat{\mathbf{y}}_{h,2}, \dots, \hat{\mathbf{y}}_{h,M_{h}}\}\) obtained from the cluster \(h\) are aligned based on word boundaries. 
The \(l\)-th word of the \(m_{h}\)-th hypothesis, \( \hat{y}_{m_{h},l} \) is aligned into a confusion network. 
For each position \( l \) in the aligned network, a voting process is applied to determine the most likely word.
The most frequent word \( w_l \) at position \( l \) is selected based on a majority vote:
\begin{align}
w_{m_{h},l} = \argmax_{w} \sum_{m_{h}=1}^{M_{h}} \mathbf{1}(\hat{y}_{m_{h},l} = w).
\end{align}
After voting for each position, the final merged transcription \( \mathbf{w}_h \) is given by:
\begin{align}
\mathbf{w}_h = \{w_{m_{h},1}, w_{m_{h},2}, \dots, w_{m_{h},l}\}.
\end{align}
This method improves overall recognition accuracy by resolving ambiguities through majority voting.

\section{Experiments}

\begin{table}[t!]
\caption{Comparison of the proposed method on clean and noisy speech (WER \%). 
HCM could not be trained stably using only 2mix data. 
Additionally, the PIT model for 3 speakers was not evaluated due to computational cost constraints.
SOT had mismatches because the training data contains time lags between utterances due to FIFO.}
  \label{table:results1}
  \centering
  \begin{tabular}{c|c|ccc|ccc}
    \hline
    \multirow{2}{*}{method} & training & \multicolumn{3}{c|}{clean} & \multicolumn{3}{c}{noisy} \\
         & data & 1mix & 2mix & 3mix & 1mix & 2mix & 3mix \\
    \hline
    PIT  &  2mix & \bf{22.5} & \bf{13.7} & \bf{53.2} & \bf{42.3} & \bf{24.4} & \bf{59.9} \\
    SOT &  2mix & 104.9 & 18.9 & 74.0 & 94.1 & 24.5 & 77.2 \\
    HCM &  2mix  & -  & - & - & -  & - & - \\
    \hline
    PIT  &  1,2mix & 7.1 & 12.5 & \bf{53.5} & 34.9  & 27.6 & \bf{62.3} \\
    SOT  &  1,2mix  & \bf{5.3} & {\bf 12.3} & 73.0 & \bf{11.8} & 22.8 & 76.7 \\
    HCM &  1,2mix  & 6.1 & 13.7 & 64.0 & 12.6 & \bf{21.4} & 74.0 \\
    \hline
    PIT  &  1,2,3mix &  - & - & - & - & - & - \\
    SOT  &  1,2,3mix  & \bf{5.4} & 18.7 & 48.0 & 11.2 & 29.3 & 57.5 \\
    HCM &  1,2,3mix  & 5.7 & \bf{8.2} & \bf{21.5} & \bf{10.4} & \bf{18.4} & \bf{36.3} \\
    \hline
  \end{tabular}
  \vspace{-10pt}
\end{table}

\begin{table}[t!]
\caption{Comparison of the merging methods on LibriMix noisy data when varying parameters related to speaker tokens (WER \%). 
}
  \label{table:results3}
  \centering
  \begin{tabular}{c|ccc|rrr}
    \hline
        merging  & \# classes & \# hypo. & candidates & 1mix & 2mix & 3mix  \\
    \hline
      \multirow{10}{*}{SIMPLE} & 32 & 4 &  random & \bf{10.2} & 35.7 & 55.6 \\
       & 32 & 4 & top-N & \bf{10.2} & 28.5 & 49.3 \\
       & 32 & 8 & top-N & \bf{10.2} & 25.6 & 46.0 \\
       & 32 & 16  & top-N & \bf{10.2} & 24.8 & 44.1 \\
       & 32 & 32  & top-N & \bf{10.2} & 24.6  & 44.0 \\
       \cline{2-7}
       & 1024 & 4 &  random & 10.4 & 31.4 & 56.2\\
       & 1024 & 4 & top-N & 10.4 & 27.5  & 53.0 \\
       & 1024 & 8 & top-N & 10.4 & 20.8 & 45.1 \\
       & 1024 & 16  & top-N & 10.4 & 19.5 & 41.8 \\
       & 1024 & 32  & top-N & 10.3 & 19.3 & 41.3 \\
    \hline
      \multirow{10}{*}{ROVER} & 32 & 4 &  random & 10.3 & 32.6 & 52.1 \\
       & 32 & 4 & top-N & \bf{10.2} & 27.8 & 47.5 \\
       & 32 & 8 & top-N & \bf{10.2} & 24.1 & 41.8 \\
       & 32 & 16  & top-N & 10.4  & 22.7 & 37.2 \\
       & 32 & 32  & top-N & 10.4  & 22.6  & 36.7 \\
       \cline{2-7}
       & 1024 & 4 &  random & 10.5 & 29.2 & 53.8\\
       & 1024 & 4 & top-N & 10.4 & 25.2 & 49.9 \\
       & 1024 & 8 & top-N & 10.5 & 19.1 & 39.4 \\
       & 1024 & 16  & top-N & 10.5 & 18.5 & 36.7 \\
       & 1024 & 32  & top-N & 10.4 & \bf{18.4} & \bf{36.3} \\
       \cline{2-7}
       & 2682 & 32 & top-N & 205.7 & 426.7 & 227.9 \\
    \hline
  \end{tabular}
    \vspace{-10pt}
\end{table}

We evaluated the proposed method using the LibriMix dataset \cite{cosentino2020librimix}. 
LibriMix contains speech superimposed on multiple speakers' utterances and noise provided by the WHAM! corpus \cite{wichern2019wham}.
For the training data, we prepared three subsets based on the number of overlapping speakers.
For the evaluation set, we prepared subsets from LibriMix with one, two, and three overlapping speakers.
Note that for SOT, the training data was simulated by delaying subsequent utterances by 1 to 1.5 seconds, which caused some mismatches.

The model for HCM is an attention-based encoder-decoder model. 
The encoder is a Conformer encoder, where we used 12 layers with 256 nodes. 
The decoder is a six-layer attention decoder. 
We performed joint training with CTC, with a CTC weight of 0.1. 
Speaker tokens were obtained using embeddings from TitaNet-large\footnote{https://catalog.ngc.nvidia.com/orgs/nvidia/teams/nemo/models/titanet\_large} \cite{koluguri2022titanet} finetuned with TSASR approach\cite{moriya22_interspeech}. 
The \(k\)-means clustering was performed with 1024 classes, which were initialized with \(k\)-means++ \cite{arthur2006k}, using VoxCeleb\cite{nagrani2020voxceleb,chung2018voxceleb2} as the training data.

For comparison, we also tested PIT and SOT. Both used a 12 layer Conformer encoder. 
In the case of PIT, the encoder was divided into four shared encoder layers and eight speaker-dependent encoder layers.
PIT also employed joint training with CTC \cite{hori2017joint}, with a CTC weight of 0.2. 
On the other hand, SOT did not use CTC joint training. 
This was because the target token sequence became considerably long, potentially causing issues with CTC calculation.

Table \ref{table:results1} presents the comparison of HCM with PIT and SOT on both clean and noisy speech data.
The table is divided into two parts: the left part shows the results for clean speech evaluation, while the right part shows the results for noisy speech evaluation. 
As shown in the table, there was a significant performance improvement, particularly in the three-speaker setting, where we achieved a 55\% relative error reduction on clean data and a 36\% relative error reduction on noisy data compared to SOT.
However, HCM faced stability issues during training when using only 2mix data, which suggests that our method requires single speaker data to get ability of speaker class estimation.
It is also important to note that the PIT model was not able to be evaluated for the three-speaker case due to computational cost constraints.
In the clean speech evaluation, HCM demonstrated significant improvements over the other methods in the scenarios with a higher number of overlapping speakers.
In the noisy speech evaluation, HCM consistently performed better than PIT and SOT, though the differences were less pronounced in the 1-mix and 2-mix conditions, indicating that noise in training data can adversely impact the effectiveness of HCM. 

We further explored the impact of varying the number of $k$-means clusters on the performance of HCM, which is shown in Table \ref{table:results3}.
The table compares different selecting strategies ("top-N" and "random") and merging strategies described in Sec. \ref{ssec:simple_voting} and Sec. \ref{ssec:HCM} where the number of clusters was set to 32 and 1024.
In the top-N condition, we selected N candidates in the order of highest probability, while in "random," we selected N candidates randomly, regardless of probability.
The results demonstrate that increasing the number of clusters generally improves the word error rate (WER), particularly in the harder 2-mix and 3-mix scenarios. 
The "top-N" approach consistently outperformed the "random" selection method, indicating that selecting the most probable speaker tokens was more effective. 
Additionally, using a larger number of hypotheses (e.g., 16 or 32) further enhanced the performance.
These findings highlight the importance of appropriately tuning the number of clusters and the merging strategy to optimize the performance of the HCM.
We also experimented with an original speaker-label-based setting, using 2,682 clusters (see the last row) where the original speaker labels were employed instead of \(k\)-means labels. 
However, this made the training unstable and performance was significantly worse.

\begin{table}[t!]
  \caption{Speaker counting accuracy (\%).}
  \label{table:results4}
  \centering
  \begin{tabular}{c|c|c|cccc}
    \hline
          & \multirow{2}{*}{eval. data} & actual & \multicolumn{4}{c}{estimated \# speakers} \\
          & & \# speakers & 1 & 2 & 3 & more\\
    \hline
      \multirow{3}{*}{SOT} & \multirow{3}{*}{clean} & 1  & 88.66 & 10.15 & 0.92 & 0.27 \\
      
        & & 2 & 0.20 & 99.5 & 0.27 & 0.00\\
        & & 3 & 0.40 & 52.80 & 41.43 & 5.37\\
    \hline
      \multirow{3}{*}{HCM}  & \multirow{3}{*}{clean} & 1  & 99.85 & 0.11 & 0.04 & 0.00\\      
        & & 2 & 0.00 & 99.90 & 0.10 & 0.00\\
        & & 3 & 0.00 & 3.27 & 91.83 & 4.90 \\
    \hline
    \hline
      \multirow{3}{*}{SOT} & \multirow{3}{*}{noisy} & 1  & 97.53 & 2.43 & 0.03 & 0.00 \\
        & & 2 & 0.53 & 98.37 &  1.00 &  0.10\\
        & & 3 & 0.90 & 47.63 & 45.33 &  6.13\\
    \hline
      \multirow{3}{*}{HCM}  & \multirow{3}{*}{noisy} & 1  & 99.40 &  0.43 &  0.03 &  1.33\\      
        & & 2 &  0.07 & 98.07 &  1.53 &  0.33 \\
        & & 3 & 0.03 &  5.20 &  77.8 & 16.97 \\
    \hline
  \end{tabular}
  \vspace{-10pt}
\end{table}


Table \ref{table:results4} presents the speaker counting accuracy for both HCM and SOT methods on clean and noisy evaluation data. 
The results indicate that HCM consistently outperformed SOT in speaker counting accuracy across all the conditions. Notably, HCM achieved nearly perfect accuracy in the 1- and 2-speaker cases, even under the noisy conditions, showcasing its robustness in a variety of environments. 
While there were slight decreases in accuracy for both methods in the 3-speaker case and noisy scenarios, HCM still maintained strong performance, demonstrating its ability to deal with more challenging conditions. 
These findings highlight the effectiveness of HCM in accurately estimating the number of speakers, even in the presence of noise; thus, our proposed HCM is a reliable method for multi-speaker recognition.

\section{Conclusion}
This paper proposes a novel approach, HCM, for overlapped speech recognition by leveraging special tokens based on speaker class IDs within an attention-based encoder-decoder framework.
HCM generates recognition results for each speaker class candidate and merges them based on the text contents. 
In the experiments on the LibriMix dataset, HCM demonstrated superior performance over traditional PIT and SOT approaches in the target-speaker-free MT-ASR scenarios. 
Key findings include the appropriate number of \(k\)-means clusters and the effectiveness of the "top-N" merging strategy with AHC+ROVER. 
These results offering a novel and promising direction for future advancements in overlapped speech recognition.

\newpage

\bibliographystyle{IEEEtran}
\bibliography{IEEEabrv,mybib}

\end{document}